\documentclass[10pt,letterpaper]{article}

\usepackage{cvpr}              

\newcommand{\cmark}{\ding{51}}%
\newcommand{\xmark}{\ding{55}}%
\newcommand{\thename}[0]{RAD-2}

\newcommand{\tablestyle}[2]{\setlength{\tabcolsep}{#1}\renewcommand{\arraystretch}{#2}\centering\footnotesize}
\usepackage{pifont}
\usepackage{multirow}
\usepackage[table]{xcolor}
\usepackage{makecell}
\usepackage{bbm}
\usepackage{marvosym} 
\usepackage{array}
\usepackage{makecell}
\usepackage{colortbl}
\usepackage{sectsty}
\usepackage{enumitem}
\usepackage{graphicx}

\usepackage{float} 
\usepackage{titlesec}
\usepackage{algorithm}
\usepackage{algorithmic}

\titlespacing*{\section}
{0pt}
{10pt plus 2pt minus 2pt}  
{6pt plus 2pt minus 2pt}   

\titlespacing*{\subsection}
{0pt}
{8pt plus 2pt minus 2pt}
{4pt plus 2pt minus 2pt}

\definecolor{horizonblue}{RGB}{0, 102, 204}

\definecolor{cvprblue}{rgb}{0.21,0.49,0.74}
\definecolor{ourlightblue}{RGB}{245,247,255}
\usepackage[pagebackref,breaklinks,colorlinks,allcolors=horizonblue]{hyperref}
\def\thename{ReDrive}

\def\paperID{*****} 
\def\confName{CVPR}
\def\confYear{2026}

\definecolor{horizonblue}{RGB}{0, 102, 204}

\title{
    \vspace{-4em} 
    \noindent
    \makebox[\textwidth][s]{ 
        \includegraphics[height=2em]{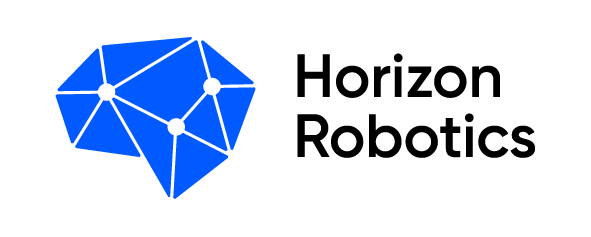} 
        \hspace{-0.5em}
        \raisebox{0.3em}{\color{gray!50}\rule{0.5pt}{1.2em}} \hspace{0.5em} 
        \includegraphics[height=2em]{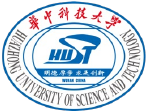} %
        \hfill 
    }
    \\
    \vspace{0.4em}
    {\color{horizonblue}\hrule height 1pt} 
    \vspace{1.0em}  
    \thename{}: Shaping Representations with \\ World Modeling for End-to-End Driving
    \vspace{1.0em} \\
    {\color{horizonblue}\hrule height 1pt} 

}

\author{
\textbf{Yueting Zhu}$^{1}$ \quad
\textbf{Shaoyu Chen}$^{2}$ \quad  
\textbf{Yuehao Song}$^{1}$ \quad
\textbf{Hui Sun}$^{2}$ \quad \\
\textbf{Qian Zhang}$^{2}$ \quad 
\textbf{Wenyu Liu}$^{1}$ \quad 
\textbf{Xinggang Wang}$^{1,\textrm{\Letter}}$ \\
\textsuperscript{1}\,Huazhong University of Science \& Technology \quad
\textsuperscript{2}\,Horizon Robotics 
}

\begin{document}
\maketitle

{
\renewcommand{\thefootnote}{}
\footnotetext{$^\textrm{\Letter}$ Corresponding author.}
}

\begin{abstract}
Driving policies require capabilities of scene understanding and future evolution prediction.
To achieve this goal, current end-to-end models typically construct complex perception-planning pipelines or introduce world models that explicitly predict future states, resulting in a complex system architecture.
Inspired by the transferability of general-purpose visual representations, we argue that combining sufficiently strong visual representations with representation world modeling can support effective planning without relying on complex inference-time auxiliary modules.
Based on this insight, we present \thename{}, an end-to-end driving framework that strengthens planning-oriented visual features via future representation prediction.
To achieve this, \thename{} adopts a three-stage training pipeline consisting of driving video pretraining, joint world-modeling and planning training, and planner adaptation.
This yields a strong planning-oriented representation and a high-performance planner, while requiring neither auxiliary perception modules nor future prediction at inference time.
Experiments on NAVSIM demonstrate strong performance, achieving 91.0 PDMS on NAVSIM v1 and 90.8 EPDMS on NAVSIM v2.
These results show that shaping representations with world modeling is sufficient to enable high-performance end-to-end planning while retaining a simple encoder-planner inference pipeline.

\end{abstract}
\vspace{-12pt}    
\section{Introduction}

End-to-end autonomous driving~\citep{uniad,vad} requires planning-oriented visual representations that capture scene semantics while reflecting future scene evolution.
Scene semantics~\citep{bevformer,drivejepa} provide the essential context for trajectory planning. 
Future scene evolution~\citep{drivelaw,latent-wam} provides predictive information for guiding future driving behaviors.

Prior work has explored diverse strategies for learning visual representations for planning.
BEV-based methods~\citep{bevformer,uniad,vad} leverage task-specific supervision to learn spatially structured BEV features.
Inspired by the transferability of self-supervised visual representations trained from large-scale visual data without task-specific annotations~\citep{dinov2,dinov3,vjepa,vjepa2}, recent works transfer these general-purpose representations to driving policies to provide strong visual priors for downstream planning~\citep{drivejepa,clear,drift}.
However, when used alone without auxiliary perception tasks, these general-purpose representations still underperform task-specific BEV representations in downstream planning.

World models provide another paradigm by explicitly modeling future scene evolution. 
Generative world models~\citep{drivelaw,epona,metis} predict future observations in pixel space, capturing scene evolution through visual generation.
However, modeling fine-grained future geometric details may distract from the scene dynamics most relevant to planning.
Recent approaches model dynamics in latent space by predicting future representations, avoiding explicit pixel generation while focusing on higher-level scene evolution~\citep{dawn,dawam}.
These methods achieve planning performance comparable to perception-based approaches without auxiliary perception tasks.
However, inference-time reliance on future representations introduces additional computational cost and pipeline complexity.

\begin{figure}[t]
    \centering

    \begin{subfigure}[t]{0.658\textwidth}
        \centering
        \includegraphics[width=\linewidth]{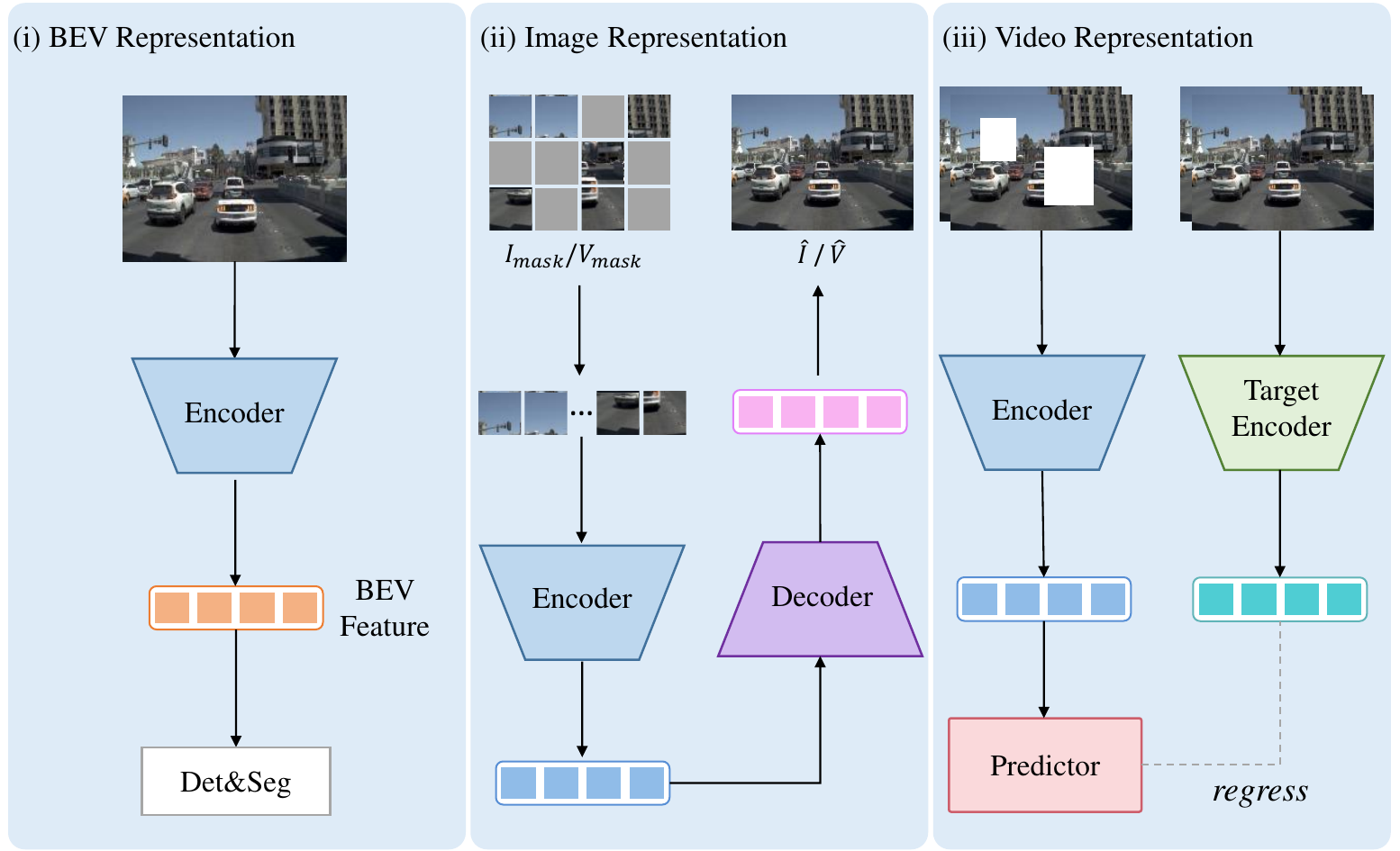}
        \caption{Existing visual representations.}
        \label{fig:intro_existing}
    \end{subfigure}%
    \begin{subfigure}[t]{0.342\textwidth}
        \centering
        \includegraphics[width=\linewidth]{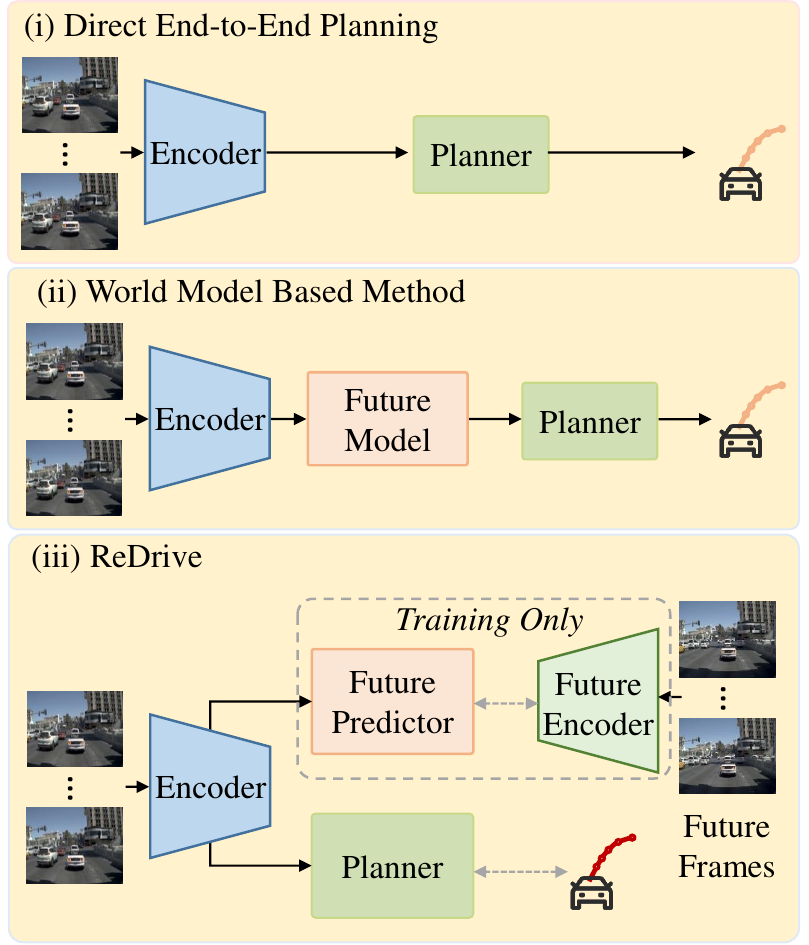}
        \caption{Comparison to existing methods.}
        \label{fig:intro_ours}
    \end{subfigure}

    \vspace{-0.5em}\caption{Overview of existing visual representation learning paradigms and our proposed \thename{} for end-to-end autonomous driving.}
    \label{fig:intro}
\end{figure}

Building on these observations, we argue that strong pretrained video representations combined with representation world modeling can support high-performance driving without complex auxiliary modules or inference-time future prediction.
We propose \thename{}, an end-to-end driving framework that strengthens planning-oriented visual representations through trajectory-conditioned future prediction.
\thename{} adopts a two-branch architecture that predicts ego trajectories and future representations conditioned on the ego trajectories. 
Jointly optimizing the two branches provides the visual encoder with additional supervision on planning-relevant scene evolution beyond direct trajectory supervision.
At inference, \thename{} uses only the learned visual encoder and trajectory generator, with the future predictor completely removed.

\thename{} follows a three-stage training procedure consisting of drive video pretraining, joint training, and planner adaptation.
We first initialize the video encoder with a general-purpose video pretrained model~\citep{vjepa2} and further pretrain it on large-scale driving videos through masked visual modeling in latent feature space.
We then jointly train the encoder with a trajectory generator and a future representation predictor.
The trajectory generator learns to predict the ground-truth ego trajectory, while the future predictor uses the expert trajectory as a condition to predict the corresponding future representation.
The shared encoder therefore receives supervision from both tasks, encouraging it to capture information about future scene evolution beyond motion supervision alone.
Finally, we freeze the visual encoder and future predictor while adapting the planner with its on-policy rollout.
We obtain the corresponding future representation of the rollout trajectory using the frozen future predictor.
Therefore, we directly optimize the planner under supervision of the future representation modeling.

We evaluate \thename{} on both NAVSIM v1~\citep{navsim} and NAVSIM v2~\citep{navsimv2}.
After training, the learned planning-oriented visual representations exhibit stronger planning capability, yielding a 3.5-point PDMS improvement over representations learned without our complete training pipeline.
\thename{} demonstrates strong trajectory planning performance across both benchmarks, achieving a PDMS of 91.0 on NAVSIM v1 and an EPDMS of 90.8 on NAVSIM v2.

Our contributions are summarized as follows:

\begin{itemize}
\item We introduce \thename{}, a simple yet effective driving policy built upon a strong planning-oriented visual representation shaped with world modeling without the requirement of inference-time auxiliary architecture designs, e.g., perception heads and future predictors.

\item We develop a three-stage training strategy that progressively incorporates future prediction supervision into trajectory learning and further optimizes the planner on its own rollouts under future representation modeling supervision.

\item We evaluate \thename{} on the NAVSIM v1 and NAVSIM v2 benchmarks, demonstrating strong planning performance across both benchmarks.

\end{itemize}

\section{Related Work}
\subsection{End-to-end Autonomous Driving}
End-to-end autonomous driving aims to directly optimize driving decisions from sensor observations.
Early approaches~\citep{st-p3,uniad,vad} build scene representations and jointly optimize perception, prediction, and planning to support end-to-end trajectory generation.
Subsequent methods~\citep{diffusiondrive,resad,diffusiondrivev2,goalflow} adopt generative planners that directly model the trajectory distribution.
Recent methods incorporate reinforcement learning~\citep{rad,rad2,alphadrive} or vision language model (VLM) prior~\citep{senna,senna2,recogdrive} to enhance planning performance.

\subsection{Driving Representation Learning}
Classical autonomous driving systems commonly build bird's-eye-view (BEV) representations from multi-view camera inputs to provide spatially structured features for downstream perception and planning~\citep{bevformer,uniad,vad}.
Recent advances in self-supervised visual pretraining offer a different paradigm.
Several works~\citep{latent-wam,drivor} have explored general-purpose image representations, such as DINOv2 features~\citep{dinov2}, as visual backbones for end-to-end driving.
More recently, a series of works~\citep{drift,drivejepa,clear} further incorporate pretrained video representations~\citep{vjepa,vjepa2} into driving scene understanding for better temporal semantic extraction.
Another direction predicts compact action-oriented representations for driving decisions~\citep{autojepa}.

\subsection{Driving World Action Models}
World action models leverage explicit scene evolution prediction to enhance trajectory generation.
Existing methods typically forecast future states in BEV or latent representations and use them for trajectory generation, candidate selection, or online evaluation~\citep{wote,world4drive,drivefuture,seerdrive}.
Other approaches connect world representations with planning policies through future visual generation~\citep{epona,drivelaw,reworld}.
These methods mainly use predicted futures or world representations as additional information for planning, making trajectory decisions directly dependent on future prediction. In contrast, we use future dynamics to improve the video representation used for planning, enabling direct trajectory planning with a stronger representation.
\section{Method}
\subsection{Overview}

\begin{figure}[t]
    \centering
    \includegraphics[width=\textwidth]{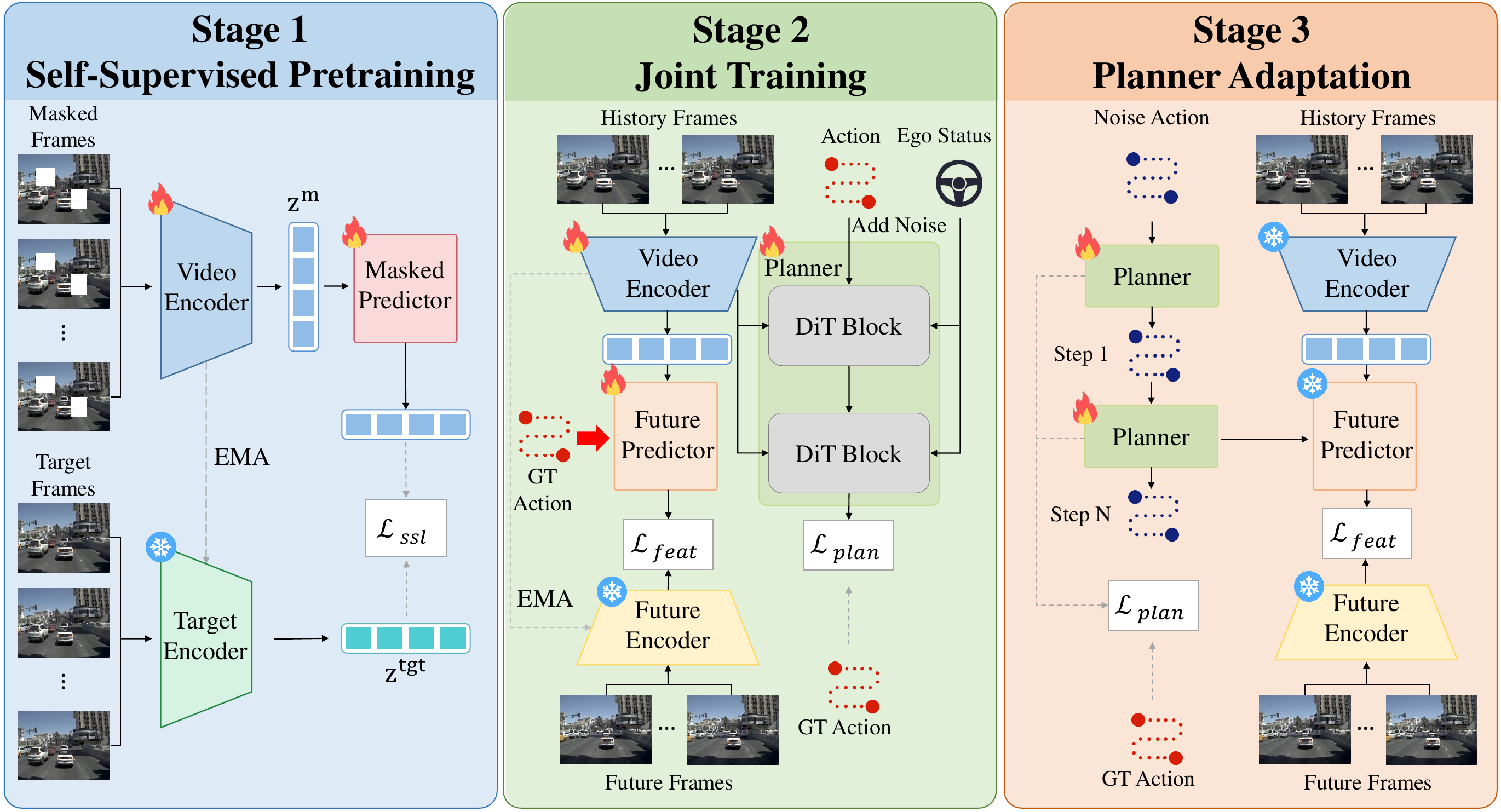}
    \caption{Overview of \thename{} and its three-stage training procedure. Self-Supervised Pretraining adapts the video encoder to the driving domain. Joint Training learns trajectory generation and future representation prediction from the shared history representation. Planner Adaptation freezes the encoder and future predictor and uses planner-generated trajectories as the condition for predictive supervision.}

    \label{fig:overview}
\end{figure}

Our goal is to demonstrate that strong planning-oriented visual representations alone can support high-performance driving planning. Such representations should capture not only the current driving scene but also its potential future evolution.
Accordingly, \thename{} introduces future representation prediction as a training signal and progressively incorporates it through three stages: Driving-Domain Pretraining, Joint Training, and Planner Adaptation, as illustrated in Fig.~\ref{fig:overview}.

\subsection{Driving-Domain Pretraining}
To provide a driving-specific temporal representation for subsequent planning-oriented learning, we first perform self-supervised pretraining on driving videos.
We initialize the encoder from a video-pretrained V-JEPA2~\citep{vjepa2} backbone and further adapt it to the driving domain.
We perform latent masked visual modeling style representation learning~\citep{vjepa} to implement continual pretraining.
Specifically, we mask a set of spatiotemporal tokens $\mathcal{M}$ and predicts their latent representations from the visible context.
We utilize a context encoder to extract representations $z$ from visible tokens, while the EMA encoder extracts target representations $z^{\mathrm{tgt}}$ from the unmasked contents.
We project $z$ to predictive representation $\hat{z}$ using a representation predictor to predict the target representations of the masked regions from the visible context.
The pretraining objective is defined as
\begin{equation}\label{eq:vjepa_pretrain}
    \mathcal{L}_{\mathrm{ssl}}
    =
    \frac{1}{|\mathcal{M}|}
    \sum_{i \in \mathcal{M}}
    \left\|
        \hat{z}_i
        -
        z_i^{\mathrm{tgt}}
    \right\|_1,
\end{equation}
where $\hat{z}_i$ and $z_i^{\mathrm{tgt}}$ denote the predicted and target representations at masked position $i$, respectively, and the $\ell_1$ distance is averaged over the feature dimension.
By optimizing this masked prediction objective on driving videos, the pretrained video representation is adapted to the driving domain.
The resulting context encoder is then used to initialize the video encoder for subsequent future feature prediction and trajectory planning.

\subsection{Joint Training}
To further shape the video representation with future-aware information relevant to planning, we jointly optimize future representation prediction and trajectory planning through a shared encoder.
We introduce two branches to perform these two tasks.
The future prediction branch learns future representations conditioned on the ground-truth ego trajectory, while the planning branch directly generates the ego trajectory from the history representation. 
This joint training introduces future-aware supervision into the video representation while keeping trajectory generation decoupled from future prediction.

\subsubsection{Future Representation Prediction}

We divide the video into a history clip $X_{1:T}$ and a non-overlapping future clip $X_{T+1:2T}$. 
The history clip is encoded by $E_\theta$ and the future clip is processed by the EMA target encoder to provide the prediction target.
We condition future feature prediction on the ground-truth ego trajectory corresponding to the future clip $X_{T+1:2T}$. 
The trajectory is encoded into action tokens $A$, which are incorporated into the predictor through cross-attention with the history representation $Z_h$ to predict the future representation,
\begin{equation}
\begin{aligned}
Z_h &= E_\theta(X_{1:T}), \qquad
Z_f^{\mathrm{tgt}} = E^{\mathrm{tgt}}(X_{T+1:2T}), \qquad
\hat{Z}_f = P_\phi(Z_h, A).
\end{aligned}
\label{eq:futurepred}
\end{equation}
We regress the predicted representation $\hat{Z}_f$ toward the stop-gradient target $Z_f^{\mathrm{tgt}}$ after feature normalization,
\begin{equation}\label{eq:l1loss}
\mathcal{L}_{\mathrm{feat}}
=
\frac{1}{N}
\left\|
\operatorname{Norm}(\hat{Z}_f)
-
\operatorname{Norm}(Z_f^{\mathrm{tgt}})
\right\|_1,
\end{equation}
where $N$ denotes the number of elements in the future representation.

The objective updates both the predictor $P_\phi$ and the video encoder $E_\theta$. 
Future representation prediction encourages the encoder to capture scene evolution conditioned on ego motion. This enriches the visual representation with future dynamics.

\subsubsection{Trajectory Generation}
The planning branch adopts an action DiT~\citep{dit} to generate trajectories directly from the video representation $Z_h$. 
During training, we construct a noisy trajectory from the ground-truth trajectory $\tau$ and Gaussian noise $\epsilon \sim \mathcal{N}(0,I)$ at noise level $t$
\begin{equation}\label{eq:linear}
\tau_t
=
(1-t)\tau + t\epsilon.
\end{equation}
We embed the noise level $t$ and the ego status $c$ into noisy action tokens $\tau_t$ via AdaLN~\citep{dit}.
The action DiT incorporates the history representation $Z_h$ through cross-attention as the visual condition and predicts the velocity $\hat{v}_t$.
Following the linear noising path in Eq.~\ref{eq:linear}, we obtain the target velocity $v_t$ and optimize the Action DiT with the flow-matching objective
\begin{equation}
\begin{aligned}
\hat{v}_t
=
G_\psi(\tau_t,t,c,Z_h),\qquad
v_t = \epsilon-\tau, \qquad
\mathcal{L}_{\mathrm{plan}}
=
\left|
\hat{v}_t-v_t
\right|_2^2,
\end{aligned}
\end{equation}
where $G_\psi$ predicts the velocity field for trajectory generation.

\subsubsection{Joint Optimization}
The future prediction and planning branches are jointly optimized through the video encoder $E_\theta$. The overall objective is
\begin{equation}
\mathcal{L}
=
\mathcal{L}_{\mathrm{plan}}
+
\lambda
\mathcal{L}_{\mathrm{feat}},
\end{equation}
where $\lambda$ controls the weight of future feature prediction loss. 
The planning objective updates $E_\theta$ and $G_\psi$, while the future prediction objective updates $E_\theta$ and $P_\phi$. 
This decoupling prevents the prediction objective from steering the trajectory policy toward actions that are easier to predict rather than better for planning. 
The action DiT is therefore optimized solely by the planning objective, while future prediction improves planning by shaping a stronger video representation.

\subsection{Planner Adaptation}
In the final stage, we further optimize the planned directly using the supervision from the future predictive information on on-policy samples.
Specifically, we use the feature prediction loss to supervise the planner through the fixed predictor, alongside the trajectory supervision.

During this stage, the video encoder $E_\theta$ and future predictor $P_\phi$ are frozen, and only the action DiT is optimized. 
Starting from Gaussian noise, the Action DiT performs a multi-step rollout to generate an ego trajectory $\hat{\tau}$. The generated trajectory is encoded into action tokens $\hat{A}$ and fed into the frozen future predictor, replacing the ground-truth trajectory condition in Eq.~\ref{eq:futurepred}. The resulting feature prediction loss $\mathcal{L}_{\mathrm{feat}}$ is propagated through the generated trajectory to optimize the planner.

For trajectory supervision, we apply the flow-matching objective at each denoising step and an L1 loss on the final generated trajectory. 
The overall adaptation objective is
\begin{equation}
\mathcal{L}_{\mathrm{plan}} = \mathcal{L}_{\mathrm{fm}} + \lambda_{\mathrm{traj}}\mathcal{L}_{\mathrm{traj}}, \qquad
\mathcal{L}_{\mathrm{adapt}} = \mathcal{L}_{\mathrm{plan}} + \lambda_{\mathrm{feat}}\mathcal{L}_{\mathrm{feat}}.
\end{equation}
The video encoder and future predictor remain frozen to preserve the future dynamics learned under ground-truth trajectory conditioning. Imperfect planner-generated trajectories could otherwise introduce erroneous supervision into the future predictor and alter the learned correspondence between ego motion and future scene evolution. Consequently, all adaptation objectives update only the Action DiT.

\section{Experiments}
\subsection{Experimental Setup}

\textbf{Datasets.}
We conduct downstream trajectory planning experiments on NAVSIM~\citep{navsim}, covering both the v1 and v2 benchmarks, including the challenging NavHard evaluation split of NAVSIM v2.
NAVSIM~\citep{navsim} provides approximately 103K training scenes and 12K test scenes for standardized planning evaluation.
For driving-domain video pretraining, we use videos from nuScenes~\citep{nuscenes} and nuPlan~\citep{nuplan}, together with an 80-hour subset of the NVIDIA PhysicalAI-Autonomous-Vehicles Dataset\footnote{https://huggingface.co/datasets/nvidia/PhysicalAI-Autonomous-Vehicles}.
These durations refer to the source collection before filtering by dataset split. Only videos from the training splits of nuScenes~\citep{nuscenes} and nuPlan~\citep{nuplan} are used for pretraining.
The pretrained video encoder is subsequently used to initialize the predictive adaptation and trajectory planning stages on NAVSIM.

\noindent\textbf{Implementation Details.}
We initialize the video encoder from the pretrained V-JEPA2~\citep{vjepa2} model and further pretrain it for 100 epochs on the driving videos using 16-frame clips.
We then jointly train the video encoder, future predictor, and action DiT for 35K steps on the \texttt{NAVTRAIN} split of NAVSIM~\citep{navsim}, where four observed frames are used to predict the subsequent four frames.
The planner adaptation stage further trains the action DiT for 5K steps with the video encoder and future predictor frozen.
We use five denoising steps for trajectory rollout during this stage.
The input resolution is $256\times512$. The future predictor consists of 12 Transformer blocks with 12 attention heads and a hidden dimension of 768.
All experiments run on 16 NVIDIA H20 GPUs.
More details are provided in Sec.~\ref{appendix:details} of the supplementary material.

\noindent\textbf{Evaluation Metrics.}
We evaluate planning performance on both NAVSIM v1 and NAVSIM v2. For NAVSIM v1, we report the Predictive Driver Model Score (PDMS) together with its individual components, including no-at-fault collision (NC), drivable area compliance (DAC), time-to-collision (TTC), ego progress (EP), and comfort(C). For NAVSIM v2, we report the Extended Predictive Driver Model Score (EPDMS) as the primary aggregate metric following the official evaluation protocol.
On NavHard, we report the individual metric components and per-stage scores (S.) for Stage-1 and Stage-2, together with the combined EPDMS.

\subsection{Comparison with State-of-the-Art Methods}
\textbf{NAVSIM v1.}
Tab.~\ref{tab:navsim_v1} compares \thename{} with state-of-the-art methods on NAVSIM v1.
Among perception-free approaches, \thename{} achieves the best PDMS of 91.0, outperforming the previous best ReWorld~\citep{reworld} by 0.6 points.
Notably, despite relying only on camera inputs and without explicit perception modules, \thename{} surpasses several perception-based methods, including Hydra-MDP~\citep{hydra}, DiffusionDrive~\citep{diffusiondrive}, GoalFlow~\citep{goalflow}, and DriveDPO~\citep{drivedpo} in terms of PDMS.
These results show that strong predictive visual representations can enable a direct planner to achieve competitive performance without relying on explicit perception modules.

\begin{table*}[t]
    \tablestyle{13pt}{1.}
    \centering
    \caption{Comparison with state-of-the-art methods on NAVSIM v1. 
    The best and second-best results are highlighted in bold and underlined, respectively.}
    \label{tab:navsim_v1}
    \begin{tabular}{lllcccccc}
        \toprule
        Type & Method & Inputs 
        & NC$\uparrow$ 
        & DAC$\uparrow$ 
        & EP$\uparrow$ 
        & C$\uparrow$ 
        & TTC$\uparrow$ 
        & PDMS$\uparrow$ \\
        \midrule

        \multirow{8}{*}{Perception-based}
        & Transfuser~\citep{transfuser} & C + L & 97.7 & 92.8 & 79.2 & 100 & 92.8 & 84.0 \\
        & VADv2~\citep{vadv2} & Camera & 97.2 & 89.1 & 91.6 & 100 & 76.0 & 80.9 \\
        & UniAD~\citep{uniad} & Camera & 97.8 & 91.9 & 92.9 & 100 & 78.8 & 83.4 \\
        & Hydra-MDP~\citep{hydra} & C + L                     & 98.4 & 97.7 & 85.0 & 100 & 94.5 & 89.9 \\
        & Hydra-MDP++~\citep{hydra++} & C + L                   & 97.6 & 96.0 & 80.4 & 100 & 93.1 & 86.6 \\
        & DiffusionDrive~\citep{diffusiondrive} & C + L                & 98.2 & 96.2 & 82.2 & 100 & 94.7 & 88.1 \\
        & GoalFlow~\citep{goalflow} & C + L                      & 98.4 & 98.3 & 85.0 & 100 & 94.6 & 90.3 \\
        & DriveDPO~\citep{drivedpo} & C + L                      & 98.5 & 98.1 & 84.3 & 100 & 94.8 & 90.0 \\
        & DriveSuprim~\citep{drivesuprim} & Camera                  & 98.6 & 98.6 & 91.3 & 100 & 95.5 & 89.9 \\
        
        \midrule

        \multirow{8}{*}{Perception-free}
        & LAW~\citep{law} & C + L                 & 97.4 &  93.3&  78.8&  100 &  91.9 &  83.8 \\
        & World4Drive~\citep{world4drive} & C + L & 97.4 & 94.3 & 79.9 & 100  & 92.8  & 85.1 \\
        & Epona~\citep{epona} & Camera            & 97.9 & 95.1 & 80.4 & 99.9 &  93.8 &  86.2 \\
        & Drive-JEPA~\citep{drivejepa} & Camera & 98.7 & 96.2 & 82.9 & 100 & 95.5 & 89.0 \\
        & DriveLaW~\citep{drivelaw} & Camera      & 99.0 & 97.1 & 81.3 & 100  & 96.7  & 89.1 \\
        & ReWorld~\citep{reworld} & Camera        & 99.1 & 98.2 & 82.0 & 99.8 &  97.7 &  \underline{90.4} \\
        & DAWN~\citep{dawn} & Camera              & 98.7 & 95.9 & 84.3 & 100  & 96.0  & 89.1 \\
        
        & \textbf{\thename{} (Ours)} & Camera & 99.1 & 97.9 & 84.3 & 100 & 97.2 & \textbf{91.0} \\
        \bottomrule
    \end{tabular}
\end{table*}

\noindent\textbf{NAVSIM v2.}
Tab.~\ref{tab:navsim_v2} compares \thename{} with state-of-the-art methods on NAVSIM v2.
\thename{} achieves the best overall performance and outperforms the second-best SparseDriveV2~\citep{sparsedrivev2} by 0.7 points and also exceeds recent world-model and world-action methods such as Latent-WAM~\citep{latent-wam}, DriveFuture~\citep{drivefuture}, CoWorld-VLA~\citep{coworld}, and DreamerAD~\citep{dreamerad}.
These results further show that predictive visual representations can support strong planning performance on NAVSIM v2 without relying on an explicit future model during inference.

\begin{table*}[t]
    \tablestyle{10.5pt}{1.}
    \centering
    \caption{Comparison with state-of-the-art methods on NAVSIM v2.
    The best and second-best results are highlighted in bold and underlined, respectively.
    NC–EC uniformly report the corrected-evaluator metrics.}
    \label{tab:navsim_v2}
    \begin{tabular}{lccccccccccc}
        \toprule
        Method
        & NC$\uparrow$
        & DAC$\uparrow$
        & DDC$\uparrow$
        & TLC$\uparrow$
        & EP$\uparrow$
        & TTC$\uparrow$
        & LK$\uparrow$
        & HC$\uparrow$
        & EC$\uparrow$
        & EPDMS$\uparrow$ \\
        \midrule
        DiffusionDrive~\citep{diffusiondrive}
        & 98.2 & 95.9 & 99.4 & 99.8 & 87.5 & 97.3 & 96.8 & 98.3 & 87.7  & 84.5 \\

        DiffusionDriveV2~\citep{diffusiondrivev2}
        & 97.7 & 96.6 & 99.2 & 99.8 & 88.9 & 97.2 & 96.0 & 97.8 & 91.0  & 87.5 \\

        WAM-Diff~\citep{wam-diff} & 99.0 & 98.4 & 99.3 & 99.9 & 87.0 & 98.6 & 96.2 & 98.1 & 78.5  & 89.7 \\

        DreamerAD~\citep{dreamerad}
        & 98.0 & 97.2 & 99.5 & 99.8 & 87.8 & 97.4 & 97.5 & 98.3 & 72.4 & 87.7 \\

        Latent-WAM~\citep{latent-wam}
        & 98.1 & 97.3 & 99.6 & 99.8 & 87.7 & 97.3 & 97.6 & 98.1 & 87.3  & 89.3 \\

        DriveFuture~\citep{drivefuture}
        & 98.8 & 99.1 & 99.6 & 99.9 & 86.6 & 98.4 & 96.4 & 98.3 & 74.8 & 89.9 \\

        CoWorld-VLA~\citep{coworld} & 99.1 & 97.0 & 99.6 & 99.9 & 87.9 & 98.5 & 97.7 & 98.2 & 86.2  & 90.0 \\

        SparseDriveV2~\citep{sparsedrivev2} & 98.1 & 98.1 & 99.6 & 99.8 & 91.1 & 97.3 & 96.9 & 98.2 & 78.4 & \underline{90.1} \\

        \midrule
         \textbf{\thename{} (Ours)}
        & 99.1 & 97.8 & 99.6 & 99.9 & 87.6 & 98.7 & 98.2 & 98.3 & 86.4
        &  \textbf{90.8} \\

        \bottomrule
    \end{tabular}
\end{table*}

\noindent\textbf{NAVSIM v2 NavHard.}
Tab.~\ref{tab:navhard} compares \thename{} with state-of-the-art methods on NAVSIM v2 NavHard.
\thename{} outperforms Metis~\citep{metis} and DiffusionDrive~\citep{diffusiondrive} in both per-stage scores and combined EPDMS.
Its strengths are most pronounced in Stage-1, where it leads the compared methods in collision avoidance, drivable-area compliance, driving-direction compliance, TTC, and lane keeping.
In Stage-2, \thename{} improves collision avoidance and drivable-area compliance over Metis~\citep{metis}.
This pattern suggests stronger safety and road compliance, with room to improve motion comfort under the more challenging evaluation conditions.

\begin{table*}[t]
\tablestyle{8.2pt}{1.}
\centering
\caption{Comparison with state-of-the-art methods on NAVSIM v2 NavHard.
S1 and S2 denote Stage-1 and Stage-2, respectively.
S. denotes the per-stage score, and EPDMS reports the combined score.
The best and second-best results are highlighted in bold and underlined, respectively.}
\label{tab:navhard}
\begin{tabular}{lcccccccccccc}
\toprule
Method & Stage
& NC$\uparrow$ & DAC$\uparrow$ & DDC$\uparrow$ & TLC$\uparrow$
& EP$\uparrow$ & TTC$\uparrow$ & LK$\uparrow$ & HC$\uparrow$
& EC$\uparrow$ & S.$\uparrow$ & EPDMS$\uparrow$ \\
\midrule

\multirow{2}{*}{LTF~\citep{transfuser}}
& S1 & 97.3 & 80.2 & 97.8 & 99.3 & 83.4 & 96.2 & 92.9 & 97.8 & 71.1 & 61.3 & \multirow{2}{*}{24.4} \\
& S2 & 79.4 & 69.0 & 85.6 & 98.5 & 83.8 & 76.7 & 47.9 & 97.0 & 70.6 & 39.2 & \\
\midrule

\multirow{2}{*}{DiffusionDrive~\citep{diffusiondrive}}
& S1 & 96.8 & 86.0 & 98.8 & 99.3 & 84.0 & 95.8 & 96.7 & 97.6 & 79.6 & 66.7 & \multirow{2}{*}{27.5} \\
& S2 & 80.1 & 72.8 & 84.4 & 98.4 & 85.9 & 76.6 & 46.4 & 96.3 & 72.8 & 40.5 & \\
\midrule

\multirow{2}{*}{GTRS-DP~\citep{gtrs}}
& S1 & 94.7 & 78.8 & 96.1 & 99.5 & 83.0 & 94.4 & 92.0 & 97.5 & 72.8 & -- & \multirow{2}{*}{23.8} \\
& S2 & 80.3 & 74.4 & 84.9 & 98.0 & 81.9 & 78.8 & 45.4 & 96.7 & 70.1 & -- & \\
\midrule

\multirow{2}{*}{GuideFlow~\citep{guideflow}}
& S1 & 96.6 & 80.5 & 96.3 & 99.3 & 82.3 & 94.9 & 91.5 & 97.7 & 67.8 & -- & \multirow{2}{*}{27.1} \\
& S2 & 87.3 & 76.7 & 88.8 & 99.2 & 84.3 & 85.1 & 49.7 & 93.1 & 44.5 & -- & \\
\midrule

\multirow{2}{*}{ReCogDrive~\citep{recogdrive}}
& S1 & 96.4 & 78.9 & 98.7 & 99.8 & 82.6 & 95.6 & 94.4 & 97.6 & 74.2 & 67.7 & \multirow{2}{*}{25.7} \\
& S2 & 80.2 & 65.0 & 82.4 & 98.7 & 85.2 & 76.9 & 43.8 & 96.6 & 71.8 & 37.6 & \\
\midrule

\multirow{2}{*}{SGDrive~\citep{sgdrive}}
& S1 & 95.8 & 87.6 & 97.8 & 99.8 & 84.4 & 94.7 & 92.9 & 97.8 & 28.9 & 71.1 & \multirow{2}{*}{25.5} \\
& S2 & 79.4 & 65.4 & 79.1 & 98.9 & 88.9 & 75.3 & 42.7 & 96.4 & 29.6 & 35.2 & \\
\midrule

\multirow{2}{*}{Metis~\citep{metis}}
& S1 & 96.6 & 87.8 & 99.0 & 99.3 & 84.5 & 95.6 & 97.8 & 97.8 & 77.8 & \underline{75.8} & \multirow{2}{*}{\underline{32.2}} \\
& S2 & 79.6 & 73.3 & 84.9 & 97.8 & 85.8 & 76.6 & 47.7 & 95.4 & 75.3 & \underline{41.7} & \\
\midrule

\multirow{2}{*}{\textbf{\thename{} (Ours)}}
& S1 & 97.9 & 93.6 & 99.4 & 99.6 & 84.2
& 96.9 & 97.8 & 97.8 & 73.3 & \textbf{82.3}
& \multirow{2}{*}{\textbf{34.4}} \\
& S2 & 82.2 & 75.8 & 84.3 & 98.2 & 87.3
& 77.9 & 47.8 & 95.9 & 54.2 & \textbf{42.3} & \\

\bottomrule
\end{tabular}
\end{table*}

\subsection{Ablation Studies}
\textbf{Effect of Visual Pretraining.}
We compare different pretraining strategies and temporal lengths in Tab.~\ref{tab:ablation_pretraining}. 
For a fair comparison, DINOv2~\citep{dinov2}, MAE~\citep{mae}, and our Stage~1 pretraining use encoders with comparable parameter counts initialized from their respective pretrained checkpoints. All encoders are further pretrained on the same driving data for the same number of epochs and evaluated after the same Stage~2 training.
DINOv2 outperforms MAE, suggesting that stronger visual representations benefit trajectory planning. Our video-based pretraining further improves over both image-based baselines, demonstrating the additional benefit of temporal representation learning. Increasing the number of frames consistently improves performance, further highlighting the importance of temporal context for planning.

\begin{table}[t]
    \centering
    \begin{minipage}[t]{0.48\textwidth}
        \centering
        \tablestyle{7.2pt}{1.1}
        \caption{Ablation on visual pretraining.
        We compare image- and video-based pretraining and study
        the temporal horizon used in pretraining.
        All results are evaluated after Stage~2 training.}
        \label{tab:ablation_pretraining}
        \begin{tabular}{lcccccc}
            \toprule
            Pretraining & DINOv2 & MAE & \multicolumn{4}{c}{V-JEPA2} \\
            \cmidrule(lr){2-2}\cmidrule(lr){3-3}\cmidrule(lr){4-7}
            Frames & -- & -- & 4 & 8 & 12 & 16 \\
            \midrule
            PDMS$\uparrow$ & 84.4 & 83.9 & 89.9 & 90.4 & 90.7 & \textbf{90.8} \\
            \bottomrule
        \end{tabular}
    \end{minipage}
    \hfill
    \begin{minipage}[t]{0.48\textwidth}
        \centering
        \tablestyle{21pt}{1.1}
        \caption{Ablation on encoder fine-tuning and future prediction in Stage~2, with the encoder pretrained on 4-frame driving videos.}
        \label{tab:ablation_joint}
        \begin{tabular}{ccc}
            \toprule
            Encoder & Future Predictor & PDMS$\uparrow$ \\
            \midrule
            Frozen    & \xmark & 85.7 \\
            Trainable & \xmark & 89.2 \\
            Trainable & \cmark & \textbf{89.9} \\
            \bottomrule
        \end{tabular}
    \end{minipage}
\end{table}

\noindent\textbf{Effect of Encoder Adaptation and Future Prediction.}
During Stage~2, starting from the same encoder pretrained on 4-frame driving videos, we ablate whether the encoder is frozen and whether the Future Predictor is jointly trained with the planner in Tab.~\ref{tab:ablation_joint}. Fine-tuning the encoder with the planning objective improves PDMS from 85.7 to 89.2. Jointly training the Future Predictor further improves PDMS to 89.9, showing that future representation prediction provides complementary supervision beyond the planning objective for learning planning-oriented visual representations.

\noindent\textbf{Effect of Training Stages.}
We evaluate each stage of the training pipeline in Tab.~\ref{tab:ablation_stage}. Starting from the original V-JEPA2 model, Stage~1 pretrains the encoder on 4-frame driving videos, improving PDMS from 88.9 to 89.2. 
Stage~2 jointly trains the Encoder, Future Predictor, and Planner, improving PDMS from 89.2 to 89.9. Stage~3 further conditions future prediction on planner-generated trajectories, improving PDMS to 90.2. These results demonstrate the complementary contributions of the three stages.

\begin{table}[t]
    \centering
    \begin{minipage}[t]{0.46\textwidth}
        \centering
        \tablestyle{16.2pt}{1.1}
        \caption{Ablation on the three-stage training pipeline. The Encoder and Planner are jointly trained for evaluation, with the Future Predictor introduced in Stage~2 and Planner-only adaptation in Stage~3.}
        \label{tab:ablation_stage}
        \begin{tabular}{cccc}
            \toprule
            Stage 1 & Stage 2 & Stage 3 & PDMS$\uparrow$ \\
            \midrule
            \xmark & \xmark & \xmark & 88.9 \\
            \cmark & \xmark & \xmark & 89.2 \\
            \cmark & \cmark & \xmark & 89.9 \\
            \cmark & \cmark & \cmark & \textbf{90.2} \\
            \bottomrule
        \end{tabular}
    \end{minipage}
    \hfill
    \begin{minipage}[t]{0.52\textwidth}
        \centering
        \tablestyle{22.0pt}{1.1}
        \caption{Planning evaluation of frozen visual encoders with newly trained planners. Stage~2 joint training is included for reference. All models use 8-frame pretraining.}
        \label{tab:frozen_encoder_probe}
\begin{tabular}{lcc}
    \toprule
    Encoder & Planner & PDMS$\uparrow$ \\
    \midrule
    8-frame Pretrained & Reinitialized & 86.7 \\
    Stage~2 Trained    & Reinitialized & 90.2 \\
    \midrule
    \multicolumn{2}{l}{Stage~2 Joint Training (reference)}
        & 90.4 \\
    \bottomrule
\end{tabular}
    \end{minipage}
\end{table}

\subsection{Experimental Analysis}
\textbf{\thename{} Indeed Improves Representations.}
We further evaluate whether Stage~2 improves the visual representation itself. With the encoder frozen and a newly initialized planner, the Stage~2 encoder achieves 90.2 PDMS, substantially outperforming the pretrained encoder at 86.7 and approaching the 90.4 of Stage~2 joint training. This shows that the improvement from Stage~2 is largely retained in the visual representation and can be directly utilized for planning.

\begin{figure}[t]
    \centering
    \includegraphics[width=\textwidth]{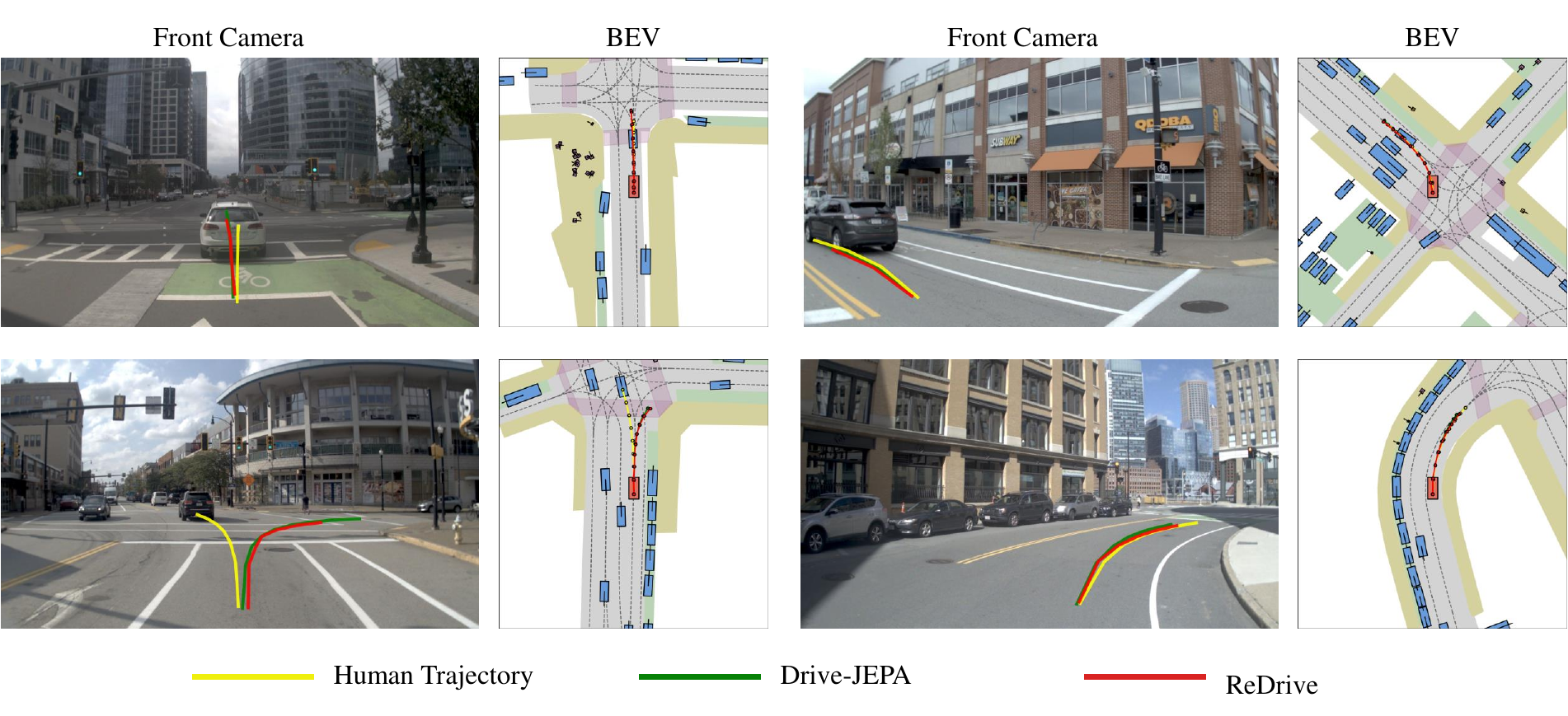}
    \caption{Qualitative comparison of planning results using front-camera and bird's-eye-view (BEV) visualizations. We compare the human trajectory, Drive-JEPA~\citep{drivejepa}, and \thename{} across representative driving scenarios.}
    \label{fig:vis}
\end{figure}

\noindent\textbf{Qualitative Results.}
Fig.~\ref{fig:vis} compares the human trajectory, Drive-JEPA~\citep{drivejepa}, and \thename{} in representative driving scenarios using both front-camera and BEV visualizations.
Compared with Drive-JEPA, \thename{} produces trajectories that more closely follow the human trajectory while maintaining better consistency with the road geometry and surrounding traffic.
The visual comparison provides qualitative evidence that the learned representation supports more accurate and scene-aware planning.
More visualizations are provided in Sec.~\ref{appendix:visualization}.

\begin{figure}[t]
    \centering
    \includegraphics[width=\textwidth]{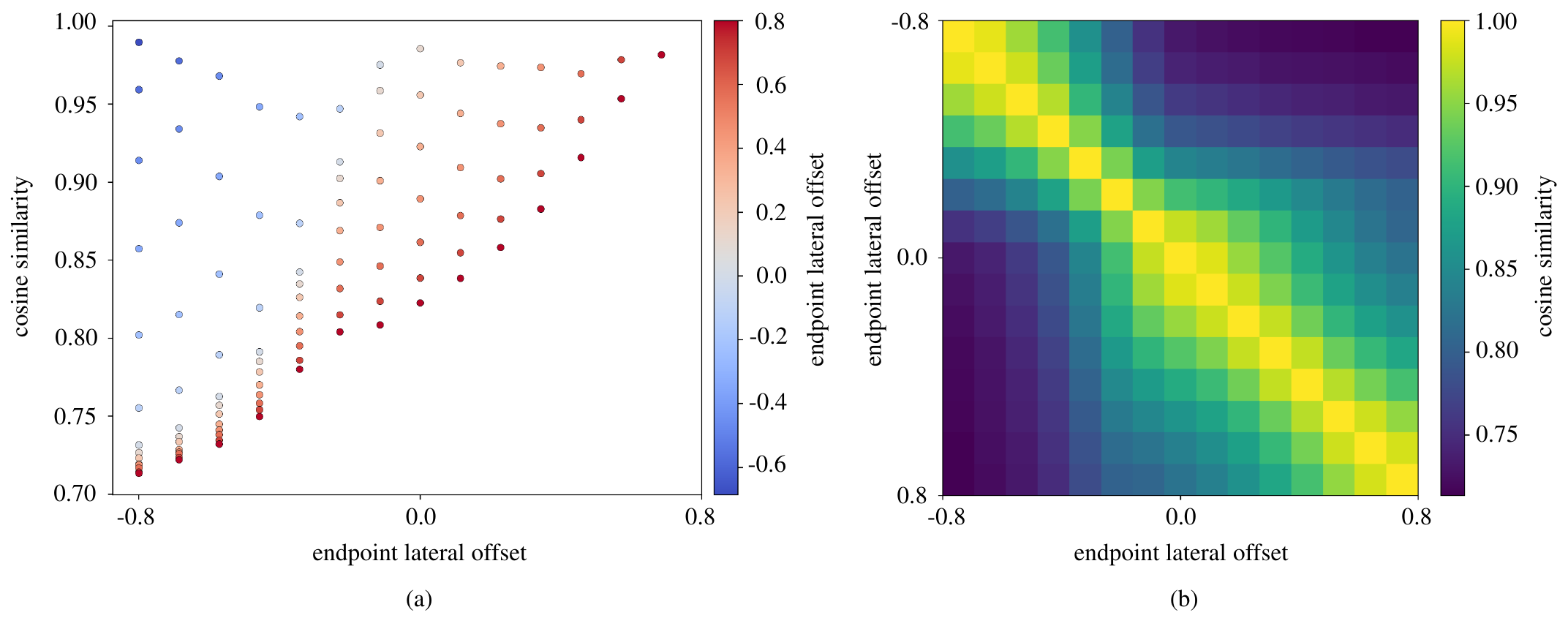}
    \caption{Sensitivity of the Future Predictor to trajectory conditions with the history representation held fixed. (a) Cosine similarities between future representations predicted under different lateral trajectory offsets. The horizontal axis and point color indicate the offsets of the two trajectory conditions being compared. (b) Pairwise cosine similarity matrix for all trajectory conditions.}
    \label{fig:analysis}
\end{figure}

\noindent\textbf{Future Predictor Analysis.}
We analyze how the Future Predictor responds to different action conditions under the same scene context.
For each scene, we keep the history representation fixed and laterally shift the ground-truth trajectory using 15 offsets uniformly spaced from $-0.8$ to $0.8$ in the normalized action space.
Each shifted trajectory is fed into the Future Predictor together with the fixed history representation to obtain the corresponding predicted future representation.
Fig.~\ref{fig:analysis}(a) plots the cosine similarities between representations predicted under different trajectory conditions, while Fig.~\ref{fig:analysis}(b) presents the corresponding pairwise similarity matrix.
The cosine similarity generally decreases as the difference between lateral offsets increases, indicating that the predicted representations are sensitive to the trajectory condition even when the scene context remains unchanged.
More visualizations are provided in Sec.~\ref{appendix:predictor}.

\section{Conclusion}
We presented \thename{}, an end-to-end driving framework that learns planning-oriented visual representations through trajectory-conditioned future representation prediction.
By jointly learning trajectory generation and action-conditioned future representation prediction, \thename{} incorporates supervision on future scene evolution into visual representation learning.
The planner is further adapted using its own generated trajectories with supervision from the frozen future predictor.
Experiments on NAVSIM v1 and v2, including NavHard, demonstrate strong planning performance.
The future predictor is removed at inference, enabling direct planning without additional modules.

{
    \small
    \bibliographystyle{ieeenat_fullname}
    \bibliography{main}
}

\clearpage
\appendix
\section{Appendix}
\subsection{Training Details}\label{appendix:details}
\subsubsection{Training Data}
\textbf{nuPlan.}
nuPlan~\citep{nuplan} is a large-scale autonomous driving dataset and planning benchmark containing approximately 1,200 hours of real-world driving data collected in Boston, Pittsburgh, Las Vegas, and Singapore.
Among them, 120 hours are released with the full sensor suite, including eight cameras, five LiDARs, an IMU, and GPS, together with detailed HD maps and automatically generated 3D annotations.
The dataset covers more than 30 types of driving scenarios, including lane changes, unprotected turns, and interactions with pedestrians.
For pretraining, we use the front-view camera sequences from the \texttt{NAVTRAIN} sensor data derived from nuPlan.

\noindent\textbf{nuScenes.}
nuScenes~\citep{nuscenes} is a large-scale multimodal autonomous driving dataset collected in Boston and Singapore.
It contains 1,000 driving scenes, each approximately 20 seconds long, with 1.4 million camera images and 390,000 LiDAR sweeps.
The sensor suite consists of six cameras, one LiDAR, five radars, an IMU, and GPS, providing complete 360-degree observations of the surrounding environment.
The official dataset is divided into 700 training scenes, 150 validation scenes, and 150 test scenes.
For pretraining, we use the \texttt{CAM\_FRONT} stream from all 700 training scenes.

\noindent\textbf{PhysicalAI-Autonomous-Vehicles Dataset.}
PhysicalAI-Autonomous-Vehicles Dataset is a large-scale multi-sensor autonomous driving dataset released by NVIDIA.
The full dataset contains approximately 1,700 hours of driving data and 306,152 clips, where each clip has a duration of 20 seconds.
The data are collected across 25 countries and more than 2,500 cities, covering diverse traffic, weather, road, and geographic conditions.
For pretraining, we use an 80-hour subset and retain the front wide-angle camera with a $120^\circ$ field of view.

\begin{figure}[!t]
    \centering
    \includegraphics[width=0.9\textwidth]{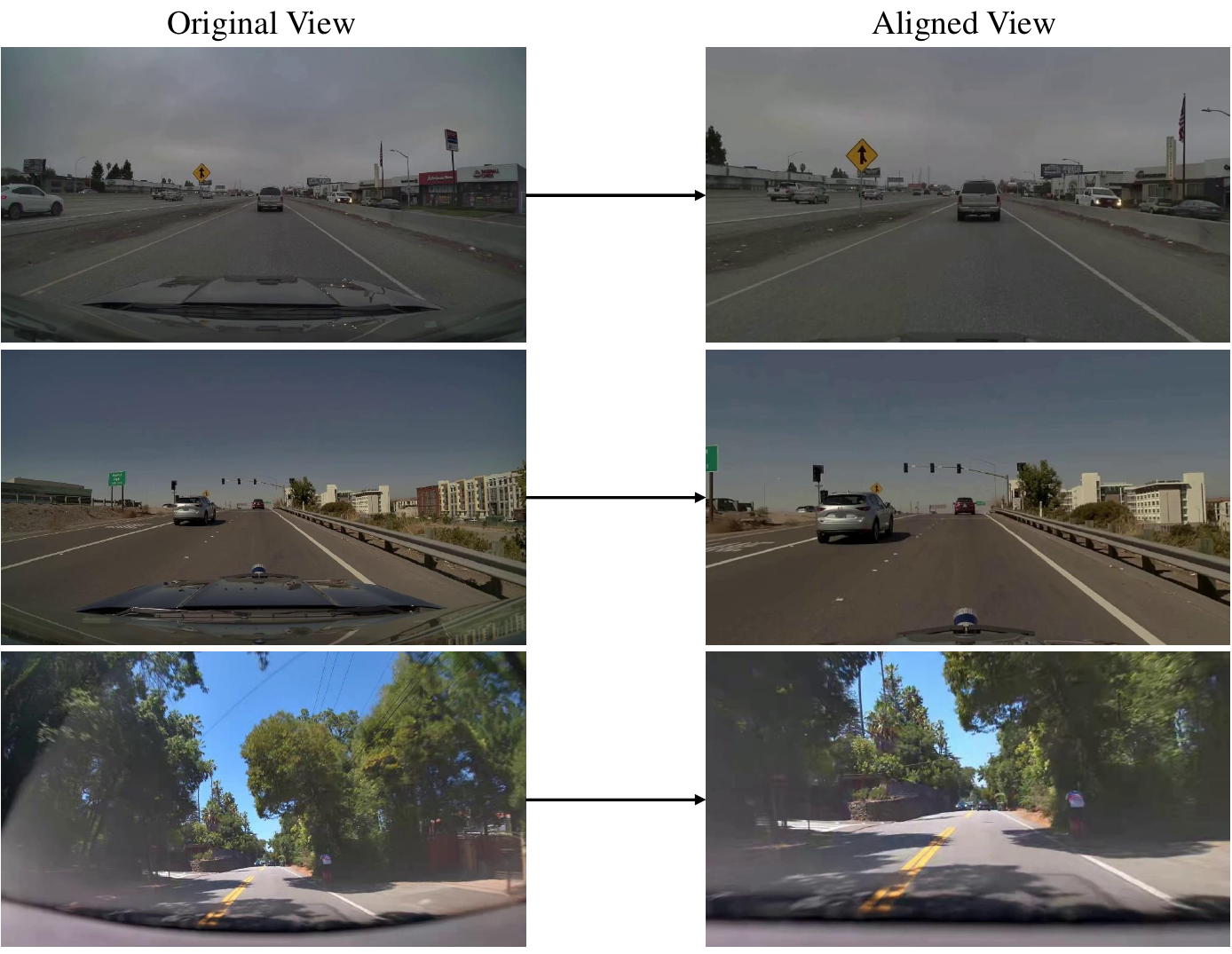}
    \caption{Illustration of camera alignment for PhysicalAI-Autonomous-Vehicles Dataset.
    The original wide-angle camera views are geometrically transformed using per-clip calibration to obtain aligned views with nuPlan-style camera geometry.
    The alignment reduces the discrepancy in camera projection and view distribution before self-supervised pretraining.}
    \label{fig:camera_alignment}
\end{figure}

\begin{figure}[!t]
    \centering
    \includegraphics[width=0.9\textwidth]{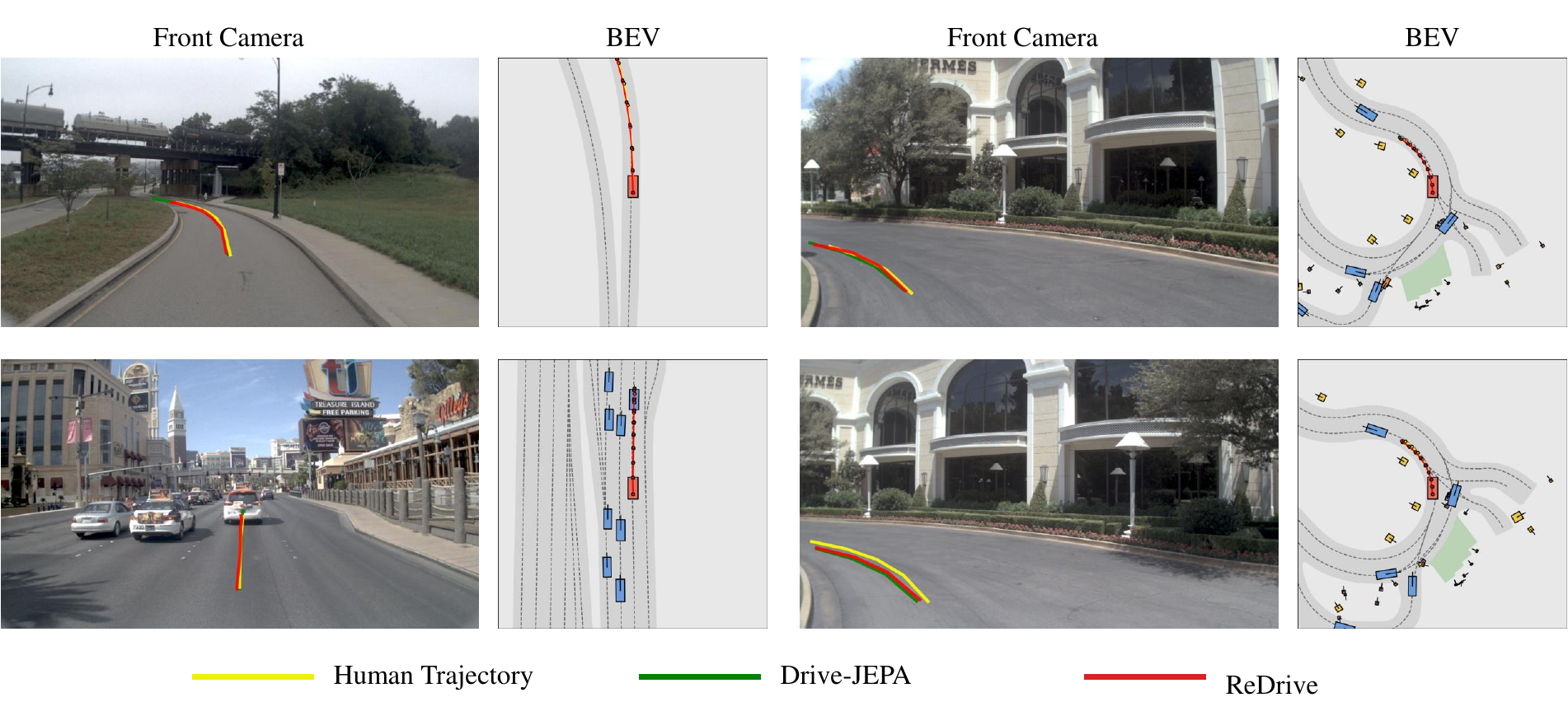}
    \caption{Additional qualitative comparison of the human trajectory, Drive-JEPA, and \thename{} using front-camera and bird's-eye-view (BEV) visualizations.}
    \label{fig:sup2_vis}
\end{figure}

\subsubsection{Camera Alignment.}
The camera configurations of the PhysicalAI-Autonomous-Vehicles Dataset
 and nuPlan~\citep{nuplan} differ substantially in field of view and projection model.
To reduce this domain discrepancy, we geometrically align the PhysicalAI-Autonomous-Vehicles Dataset front wide-angle videos to the nuPlan front-camera view before pretraining, as shown in Fig.~\ref{fig:camera_alignment}.
Specifically, we use the per-clip f-theta camera calibration provided by the PhysicalAI-Autonomous-Vehicles Dataset to map each target nuPlan pixel ray back to the corresponding source image location, followed by bilinear resampling.
The resulting videos approximately match the field of view and camera intrinsics of the nuPlan front camera.
We additionally discard clips whose calibrated field of view is insufficient to cover the target view, avoiding extrapolation during remapping.

\subsubsection{Training Procedure}
\textbf{Pretraining.}
We initialize the visual encoder from the publicly released V-JEPA2 ViT-L~\citep{vjepa2} checkpoint and further pretrain it on a mixture of driving videos from nuPlan~\citep{nuplan}, nuScenes~\citep{nuscenes}, and the PhysicalAI-Autonomous-Vehicles Dataset.
The three datasets are uniformly sampled during training.
Each training clip contains 16 frames sampled at 10\,Hz with a spatial resolution of $256\times512$.
We follow the V-JEPA2 pretraining recipe and adopt a ViT-L encoder together with a lightweight predictor and an EMA target encoder.
A multi-block masking strategy is applied over the full temporal extent, and the predictor learns to reconstruct the masked latent representations produced by the target encoder.
After pretraining, only the visual encoder is retained for subsequent training, while the pretraining predictor and target encoder are discarded.
The detailed pretraining configuration is summarized in Tab.~\ref{tab:pretraining_details}.

\begin{table}[t]
    \tablestyle{56.0pt}{1.12}
    \centering
    \caption{Pretraining configuration.}
    \label{tab:pretraining_details}
    \begin{tabular}{lll}
        \toprule
        Module & Configuration & Setting \\
        \midrule

        \multirow{5}{*}{Encoder}
        & Architecture & ViT-L \\
        & Depth & 24 \\
        & Hidden dimension & 1024 \\
        & Attention heads & 16 \\
        & Parameters & 304M \\
        
        \midrule

        \multirow{4}{*}{Predictor}
        & Depth & 12 \\
        & Hidden dimension & 384 \\
        & Attention heads & 12 \\
        & Parameters & 22M \\
        \midrule

        \multirow{4}{*}{Input}
        & Frames & 16 \\
        & Frame rate & 10\,Hz \\
        & Resolution & $256\times512$ \\
        & Patch / Tubelet size & $16\times16$ / 2 \\
        \midrule

        \multirow{3}{*}{Masking}
        & Small blocks & 8, scale 0.15 \\
        & Large blocks & 2, scale 0.70 \\
        \midrule

        \multirow{5}{*}{Optimization}
        & Optimizer & AdamW \\
        & Learning rate & $5.25\times10^{-4}$ \\
        & Weight decay & 0.04 \\
        & Precision & bfloat16 \\
        & Epochs & 100 \\
        \midrule

        \multirow{2}{*}{Target Encoder}
        & Update & EMA \\
        & Decay & 0.99925 \\

        \bottomrule
    \end{tabular}
\end{table}

\noindent\textbf{Joint Training.}
After pretraining, we retain the visual encoder and jointly optimize it with a newly initialized Future Predictor and Action DiT.
Each sample contains four observed frames at $256\times512$ resolution.
The Future Predictor takes the history representation and ego trajectory as conditions to predict the target representations of the subsequent four frames, while the Action DiT predicts an 8-step trajectory covering 4\,s at 2\,Hz.
A target encoder is maintained as an exponential moving average (EMA) of the visual encoder with a decay of 0.9999 and remains gradient-free throughout training.
The model architecture and optimization settings are summarized in Tab.~\ref{tab:joint_model_details} and Tab.~\ref{tab:joint_training_details}, respectively.

\begin{table}[t]
    \centering

    \begin{minipage}[t]{0.48\linewidth}
        \centering
        \tablestyle{18pt}{1.05}
        \caption{Model configuration for joint training.}
        \label{tab:joint_model_details}
        \begin{tabular}{lll}
            \toprule
            Module & Configuration & Setting \\
            \midrule

            \multirow{5}{*}{Encoder}
            & Architecture & ViT-L \\
            & Depth & 24 \\
            & Hidden dim. & 1024 \\
            & Attention heads & 16 \\
            & Parameters & 304M \\
            \midrule

            \multirow{6}{*}{Future Predictor}
            & Depth & 12 \\
            & Hidden dim. & 768 \\
            & Attention heads & 12 \\
            & MLP ratio & 4 \\
            & Parameters & 153M \\
            \midrule

            \multirow{5}{*}{Action DiT}
            & Depth & 28 \\
            & Hidden dim. & 512 \\
            & Attention heads & 16 \\
            & Parameters & 103M \\
            \midrule

            \multirow{2}{*}{Target Encoder}
            & Update & EMA \\
            & Decay & 0.9999 \\

            \bottomrule
        \end{tabular}
    \end{minipage}
    \hfill
    \begin{minipage}[t]{0.48\linewidth}
        \centering
        \tablestyle{27pt}{1.05}
        \caption{Optimization configuration for joint training.}
        \label{tab:joint_training_details}
        \begin{tabular}{ll}
            \toprule
            Configuration & Setting \\
            \midrule
            Observed frames & 4 \\
            Predicted frames & 4 \\
            Resolution & $256\times512$ \\
            Optimizer & AdamW \\
            Learning rate & $3\times10^{-5}$ \\
            $\beta_1,\beta_2$ & $0.9,\,0.95$ \\
            Weight decay & $1\times10^{-5}$ \\
            LR schedule & Constant w/ warmup \\
            Warmup steps & 1,000 \\
            Gradient clipping & 1.0 \\
            Planning loss & Flow-matching MSE \\
            Prediction loss & L1 \\
            Prediction weight & 0.1 \\
            Diffusion timesteps & 1,000 \\
            Inference steps & 5 \\
            Precision & bfloat16 \\
            \bottomrule
        \end{tabular}
    \end{minipage}

\end{table}

\subsection{More Qualitative Results}\label{appendix:visualization}
We provide additional qualitative comparisons between the human trajectory, Drive-JEPA~\citep{drivejepa}, and \thename{} in Fig.~\ref{fig:sup2_vis}.
The examples cover diverse road geometries and traffic interactions, with both front-camera and bird's-eye-view (BEV) visualizations.
These results further illustrate the planning behavior of \thename{} across different driving scenarios.

\subsection{Additional Future Predictor Visualizations}\label{appendix:predictor}

\begin{figure*}[!t]
    \centering
    \includegraphics[width=0.91\textwidth]{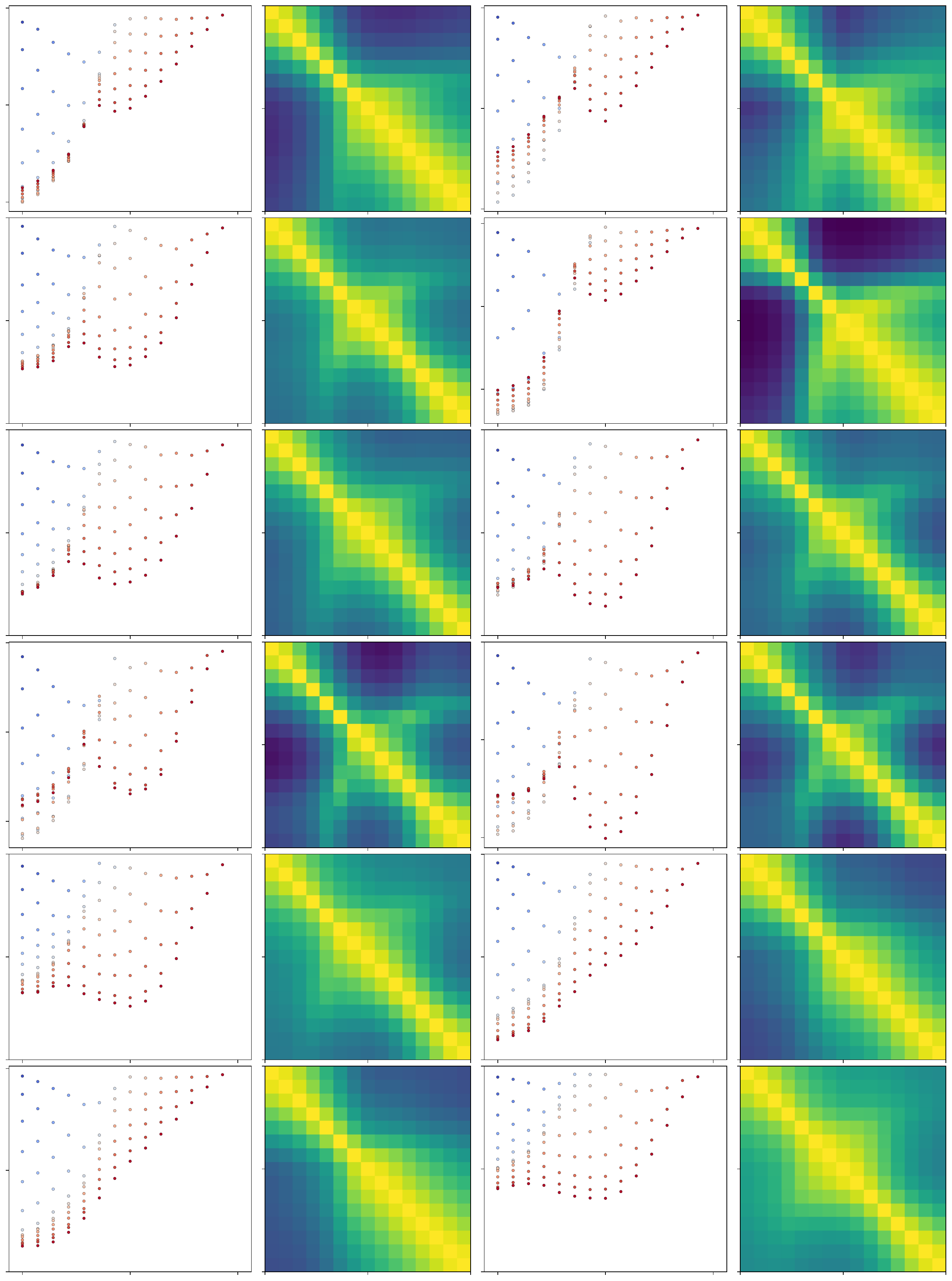}
    \caption{Additional Future Predictor visualizations.
    Predicted future representations under different lateral trajectory conditions across diverse driving scenes. The representations exhibit consistent action-dependent variations across different scenarios.}
    \label{fig:action_sweep_grid}
\end{figure*}

We provide additional visualizations of the Future Predictor across diverse driving scenes. 
We keep the history representation fixed and vary the lateral trajectory condition to examine the corresponding changes in predicted future representations. 
As shown in Fig.~\ref{fig:action_sweep_grid}, the representations consistently vary with the trajectory condition across different scenes. 
The pairwise similarity matrices also exhibit a clear decay as the difference between action conditions increases. These results further show that the Future Predictor captures action-dependent future scene evolution across diverse driving scenarios.

\end{document}